\documentclass[letterpaper, 10 pt, conference]{ieeeconf}  

\IEEEoverridecommandlockouts                              

\usepackage{graphicx} 
\usepackage{amsmath} 
\usepackage{amssymb}  
\usepackage{bm}

\title{\LARGE \bf
	Improving drone localisation around wind turbines\\ using monocular model-based tracking
}

\author{Oliver Moolan-Feroze$^{1}$, Konstantinos Karachalios$^{2}$, Dimitrios N. Nikolaidis$^{2}$, and Andrew Calway$^{1}$
\thanks{$^{1}$Oliver Moolan-Feroze and Andrew Calway are with the Department of Computer Science, University of Bristol, 75 Woodland Road, Bristol, BS8 1UB, United Kingdom
        {\tt\footnotesize oliver.moolan-feroze@bristol.ac.uk, andrew.calway@bristol.ac.uk}}%
	\thanks{$^{2}$Konstantinos Karachalios and Dimitrios N. Nikolaidis are with Perceptual Robotics, 5 Hope Road, Bristol, UK, BS3 3NZ, United Kingdom
        {\tt\footnotesize kostas@perceptual-robotics.com, dimitris@perceptual-robotics.com}}%
}

\begin{document}

\maketitle
\thispagestyle{empty}
\pagestyle{empty}

\begin{abstract}
We present a novel method of integrating image-based measurements into a drone navigation system for the automated inspection of wind turbines. We take a model-based tracking approach, where a 3D skeleton representation of the turbine is matched to the image data. Matching is based on comparing the projection of the representation to that inferred from images using a convolutional neural network. This enables us to find image correspondences using a generic turbine model that can be applied to a wide range of turbine shapes and sizes. To estimate 3D pose of the drone, we fuse the network output with GPS and IMU measurements using a pose graph optimiser. Results illustrate that the use of the image measurements significantly improves the accuracy of the localisation over that obtained using GPS and IMU alone.
\end{abstract}

\section{INTRODUCTION}
Due to harsh weather conditions, wind turbines can incur a wide range of structural damage~\cite{Ribrant2007}, which can severely impact their power generation abilities~\cite{GarciaMarquez2012}. To address this, regular inspections are needed. Current best practice in visual inspection is the use of ground-based cameras with telephoto lenses, or manual inspection using climbing equipment. Both methods incur considerable cost in both the inspection itself, and the  turbine down time. Manual inspections can also lead to inconsistencies in the data gathering, which can be compounded over multiple visits.

To address the above, inspecting using unmanned autonomous vehicles (UAVs) -- or drones -- is being considered~\cite{Wang2017}. Autonomous inspections have the potential to save time and cost and give more consistent inspection data. A key element for successful inspections is the ability to accurately determine drone location with respect to the wind turbine. Accurate localisations increase the consistencies of inspections, allow the drone to get closer to the turbine, and are useful in image post-processing. Although global positioning systems (GPS) and inertial measurement units (IMUs) can provide relatively good tracking, improved performance can be obtained by inclusion of image-based sensor readings~\cite{Leutenegger2015}. 

We present a novel system for integrating image-based measurements with GPS/IMU readings, which gives improved localisation of the drone. To do this we take a model-based tracking approach. An internal 3D skeleton representation of the wind turbine is matched to that inferred from image data using a convolutional neural network (CNN). The difference is minimised using a pose graph optimiser which is constrained by the GPS/IMU measurements.

\begin{figure}[t]
	\centering
	\includegraphics[width=1.0\linewidth]{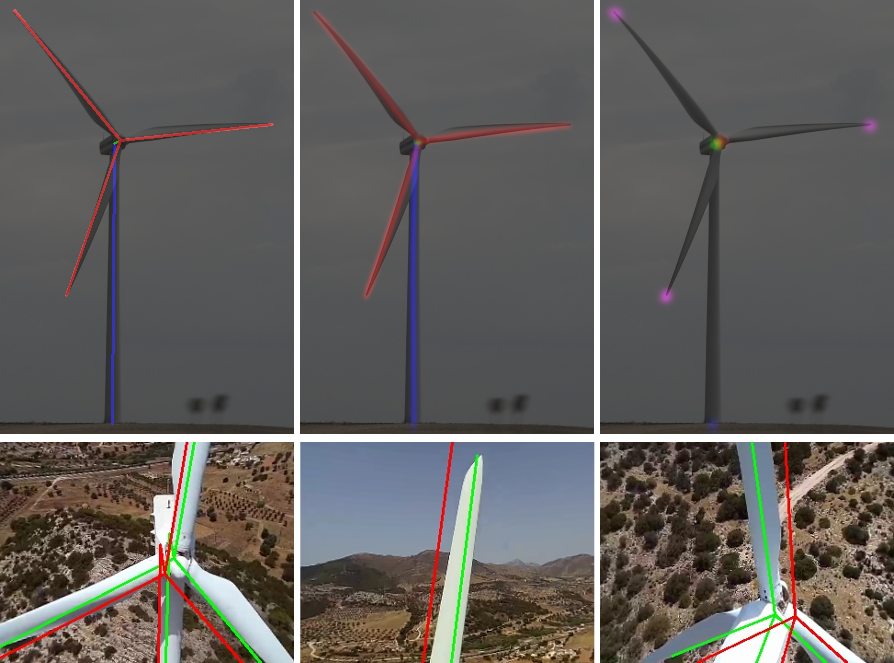}
\caption{Top Left) Wind turbine with the lines from $\mathcal{L}^c$  overlaid. Top Middle) Line output from the CNN. Top Right) Point output from the CNN. Bottom) Example of the system applied to an inspection flight. Red lines are the reprojections of the model before optimisation. Green lines are the reprojections after optimisation.}
\vspace{-4ex}
	\label{fig:model_image}
\end{figure}



There are two main contributions in this work. The first is a novel application of a CNN that is able to infer from image data the 2D projection of the 3D skeleton model. This enables us to easily find correspondences between the model and images. In addition, we incorporate prior information about the likely pose of the camera into the network to improve its prediction performance. The second contribution is the integration of the network output into a pose graph optimisation, using both point and line features.

In Section~\ref{sec:previous_work}, we explore some of the work related to the use of CNNs in localisation and tracking applications, as well as detailing the research into drone inspections of wind turbines. We present our method in Section~\ref{sec:method}, and detail the turbine representation we have chosen, give a detailed description of the CNN, and describe how the CNN outputs are integrated with the pose optimiser. In Section~\ref{sec:experiments} we describe the evaluation of our method, using both real and simulated data. Finally, in Section~\ref{sec:conclusion} we give some conclusions and ideas for future work.

\section{PREVIOUS WORK}\label{sec:previous_work}

Over recent years, deep learning has been applied to image-based camera pose estimation in a number of ways. These can be roughly separated into two groups: end-to-end approaches, and approaches that use deep learning as an intermediate or preprocessing step. The idea behind the end-to-end group is that given an input image, the 6DoF pose of the camera can be regressed directly by the network. The first attempt at this was PoseNet~\cite{kendall_posenet:_2015}, were the authors designed a VGG~\cite{simonyan_very_2014} style CNN with a 3D translation regression block and 4D quaternion regression block as outputs. Through the use of transfer learning the authors trained the network for indoor and outdoor scenes using only a small number of pose-labelled images. This work was later expanded on with the inclusion of measures of uncertainty into the output~\cite{kendall_modelling_2015} using Bayesian deep learning. Further additions were made in~\cite{kendall_geometric_2017}, including a novel geometrical loss function based on reprojection errors. Clark et.al.~\cite{clark_vidloc:_2017} take advantage of temporal smoothness in camera poses over neighbouring video frames through the use of Recurrent Neural Network (RNN) layers. A series of Long Short-term memory units are appended after a CNN block which are able to integrate features from previous time steps to aid in the regression process. This method is also used in~\cite{wang_deepvo:_2017}, where the authors state the recurrent layers are able to learn the motion dynamics of the system over a period of time.

End-to-end approaches are beneficial in their simplicity, but there are a number of problems. The most obvious is the need for pose-labelled training data. This is at best expensive and time consuming to obtain and in some cases -- such as in our application -- impossible to gather. In addition, the performance of end-to-end methods is still lagging behind traditional geometric approaches.

One way of addressing these problems is to use deep learning as an intermediate or preprocessing step in a traditional geometric approach. In~\cite{Pavlakos}, the authors use a CNN to generate heatmaps of model feature points. The location of different features are represented by peaks in the heatmaps, and the pose of the camera is obtained through a minimisation process. The authors use a stacked hourglass~\cite{newell_stacked_2016} architecture which is able to integrate features from across the spatial extent of the input image. In~\cite{Moolan-Feroze2018}, the authors propose a similar method of feature points extraction using CNNs. However, in this work they explicitly handle cases where the object is partially out-of-view. This is especially beneficial for tasks such as industrial inspection where due to close proximity, only incomplete views of the inspected object are visible.

There is not much literature related to the autonomous inspection of wind turbines using drones. Stokkeland at. al.~\cite{Stokkeland2015} present a method to determine the position of the drone and the configuration of the wind turbine during an initial approach stage. Their method uses a Hough Transform to locate the different parts of the turbine, and then integrates this information through a Kalman Filter to track the position of the drone. One drawback of this work is that it only addresses the initial approach of the drone. The actual details of the inspection are not handled. The work in~\cite{Schafer2016} propose the use of a LiDAR sensor to aid in navigation. It describes a 3D occupancy grid which is able to integrate multiple noisy sensor reading using a Bayesian update scheme. This grid then serves as a map for path planning and localisation. This work has a number of important omissions however. First, it is not applied to real data, only performed in simulation. Second, it makes no attempt at localisation, focusing only on mapping and path planning.  

\begin{figure*}[t]
	\centering
	\includegraphics[width=0.8\linewidth]{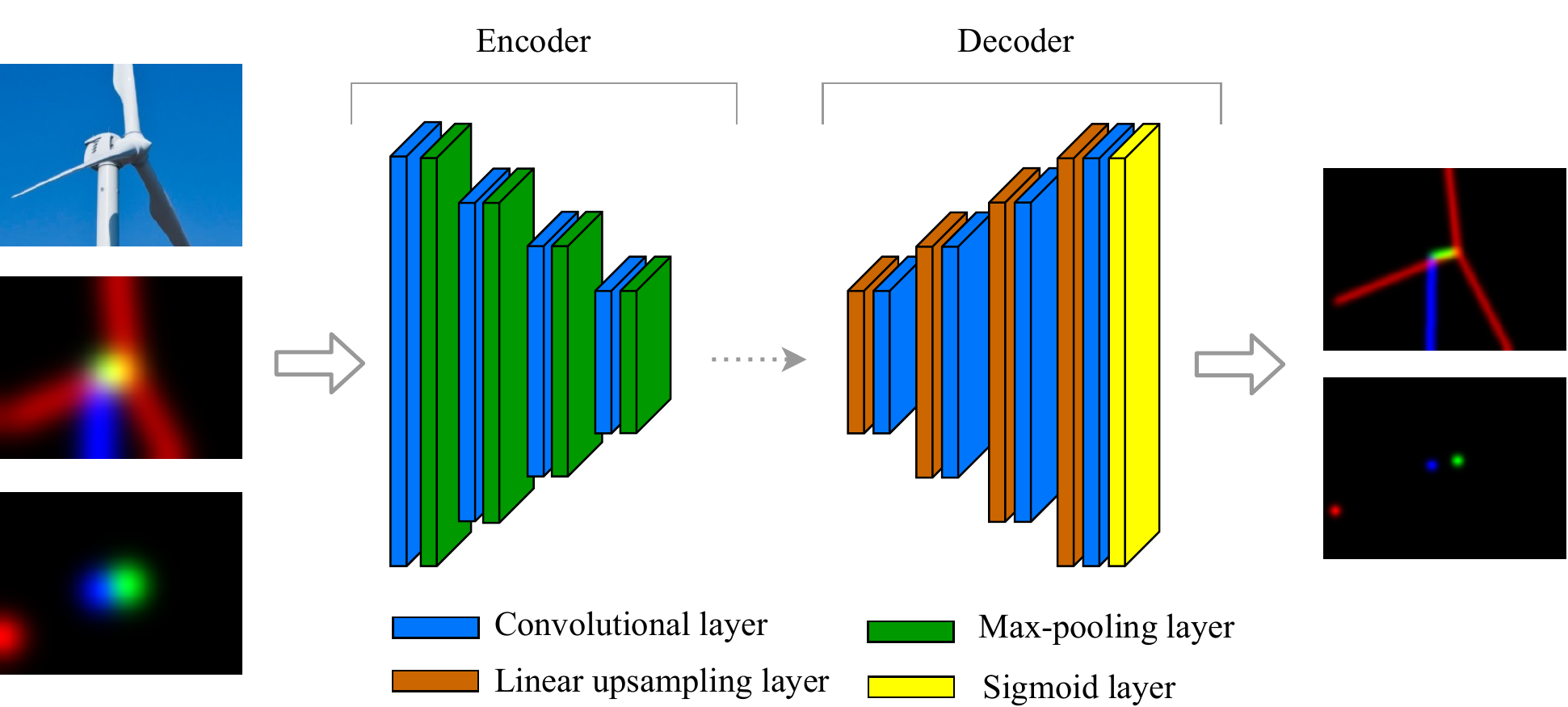}
	\vspace{-1ex}
	\caption{Overview of the CNN network showing layer architecture, inputs and outputs}
	\vspace{-1.5em}
	\label{fig:cnn_network_overview}
\end{figure*}

\section{METHOD}\label{sec:method}

The process of localising the drone with respect to the wind turbine is split into two parts. During the first part (Section~\ref{sec:cnn}), we obtain images of the turbine from the monocular camera on the front of the drone and pass them through a CNN to extract an estimate of the projection of the 3D skeleton representation. These estimates are then used to constrain a pose graph optimisation (Section\ref{sec:graph}). Key to both these parts is how the wind turbine is represented by the system (Section~\ref{sec:representation}). 

\subsection{Turbine Representation}\label{sec:representation}

To enable a model-based tracking approach, an internal representation of the wind turbine is necessary. When estimating the pose of the camera, the representation is projected through the current estimate of the pose and the camera intrinsic matrix and then compared to the visual information contained within the image. The optimal pose is then the one that aligns the reprojection of the representation with the object in the image. We have chosen a very simple skeleton representation that is general enough such that it is able to fit to a wide range of different turbine shapes, sizes and configurations.

The representation is based on a set of 3D points, combined with a set of lines which connect them. The points $\mathcal{P}^c$ lie at the base of the turbine tower, the top of the turbine tower, the centre of the blades, and the tips of each of the blades. These points were chosen by looking for the commonalities between different wind turbine shapes and sizes. The set of lines $\mathcal{L}^c$  connect the bottom and the top of the tower, the top of the tower and the centre of the blades, and the centre of the blades with each of the blade tips. In total, this gives us a set of 6 points and 5 lines. An example of the representation can be seen in Figure~\ref{fig:model_image}. We assume that a good estimate of the size and shape of the turbine is known prior to applying the localisation process.

\subsection{Model projection inference using CNNs}\label{sec:cnn}

As described in Section~\ref{sec:representation}, the internal representation of the wind turbine is intentionally very general. However, as different wind turbines can present a wide range of visual information, it is difficult to find the accurate correspondences needed for localisation. Indeed, apart from the tips of the blades, none of the points in $\mathcal{P}^c$ correspond to specific image features that would be common across all wind turbines. Furthermore, the lines in $\mathcal{L}^c$ do not run along specific edges in the images, but rather through the centre of the different parts of the turbine. To address this, we make use of a CNN to process the input images into a form that can be easily matched to the projection of the skeleton model.

The network takes a multichannel image as an input and produces a multichannel image as an output using a convolutional-deconvolutional architecture. The input is successively convolved and downsampled using Max-Pooling up to the bottleneck of the network. Next, the data is convolved and upsampled up to its original dimensions. This type of architecture is beneficial in that visual information from across the spatial range of the input is brought together in the deepest part of the network to provide feature rich information to the output of the deconvolutional part of the architecture. An visual overview of the network can be seen in Figure~\ref{fig:cnn_network_overview}.

The role of the network is to take an image of a wind turbine as input and produce an equivalent image showing the inferred projection of the turbine skeleton model detailed in Section~\ref{sec:representation}. For the line model $\mathcal{L}^c$, this will be an image $\mathbf{I}^{L}$ containing a line running up the tower, a line connecting the tower with the blades, and a line running along the centre of each of the blades. For the point model $\mathcal{P}^c$, this will produce an image $\mathbf{I}^{P}$ with a point at the bottom of the tower, a point at the top of the tower, a point at the centre of the blades, and points at the tips of the blades. As the lines and points in the two models correspond to different parts of the wind turbine, we separate the corresponding outputs into different classes. For $\mathbf{I}^L$, the tower is one class, the connecting line another class, and the lines running along the blades a third class. We don't separate the individual blades into different classes due to their rotational symmetry. For $\mathbf{I}^P$, the classes are the tower base, the tower top, the blade centre and the blade tips. Again, all three blade tips are the same class. 

For a typical convolutional-deconvolutional network, the input would consist of just the RGB image of the turbine. However, in this work, associated with each image is an estimation of the camera pose obtained from the GPS/IMUs on the drone. We use this information to act as a prior on the line and point locations, making prediction easier for the network. To do this, we construct the skeleton model, and project it through the pose estimate and camera intrinsics to find locations of each part on the input image. As the error in the pose estimates is not excessive, these projections will lie close to their true locations. To ease the work of the network, we apply Gaussian smoothing to the projections and then append these channels onto the input image and feed it into the network. This means that the network input is a ten channel image. Three channels for the RGB image, three channels for the line model priors and four channels for the point model priors. Examples of the input and outputs can be seen in Figure~\ref{fig:cnn_network_overview}.

To train the network, we obtained a set of $\sim$1000 images of wind turbines from the internet. Each of the images in the data set was manually labelled by placing landmarks at the base of the tower, the top of the tower, the centre of the blades and the tips of the blades. These 2D locations correspond to the 3D locations of the points in $\mathcal{P}^c$. We then use these landmarks to generate the labels used during training. For the point-based label images, we set the 2D pixel location of the landmark in the correct image channel to 1, with the remaining pixels set to 0. We then apply a Gaussian smoothing kernel with $\sigma=5$ to increase the spread of the landmark in the image. Each channel of the image is then renormalised to between $0-1$. For the line-based label images, we draw lines on the images by connecting the landmarks in the same way the landmarks are connected in $\mathcal{L}^c$. Again, we apply a Gaussian kernel with $\sigma=5$ to the images to increase the spread and then renormalise.

To generate the priors, we applied a set of random affine transformations to the image landmarks. This was done to replicate the amount of error we would expect to see in the GPS/IMU pose estimates during a live flight. After the transformations, we create the images in the same way that the label images are made, but apply a larger amount of Gaussian smoothing ($\sigma=20$). For each training batch, we increased the variability in the training data through augmentation. This was done by applying random translations, rotations, scaling and cropping prior to the generation of the labels and priors.  We trained the network using the Adam optimiser with a learning rate of 0.001. Training was stopped when the test loss started to diverge from the training loss after 2 days.

\subsection{Pose Graph Optimisation}\label{sec:graph}

The problem of estimating the drone's 3D position and orientation is modelled as a pose graph. As the drone performs an inspection, at regular intervals a new node or keyframe is added to the graph. This node contains an estimate of the absolute pose $\mathbf{T}_i = \left(\mathbf{R}_i, \mathbf{t}_i\right)$ obtained from the GPS/IMU, where $\mathbf{R}_i$ is the orientation and $\mathbf{t}_i$ is the 3D position. The aim is to optimise the graph using a set of constraints, such that the poses at the nodes converge to the drone's true location and orientation $\mathbf{T}^*_i$. The graph constraints are built using the pose estimates and the inferred projection of the 3D skeleton model produced by the CNN. Similar to most pose graph methods, the absolute pose estimate is not used during optimisation. Instead, it is used to compute the relative pose offset $\mathbf{T}_{i,i-1}$ between the current keyframe $\mathbf{T}_i$ and the previous keyframe $\mathbf{T}_{i-1}$, i.e.
\begin{equation}\label{eq:relative-pose-t-offset}
	\mathbf{T}_{i,i-1} =
	\begin{bmatrix}
		\mathbf{t}_{i,i-1}\\[0.5em]
		\mathbf{q}_{i,i-1}
	\end{bmatrix}
	=
	\begin{bmatrix}
		\mathbf{R}_i^T \left(\mathbf{t}_{i-1} - \mathbf{t}_i \right) \\[0.5em]
		\mathbf{q}^{-1}_i * \mathbf{q}_{i-1}
	\end{bmatrix} \enspace,
\end{equation}
where $\mathbf{q}_i$ is the quaternion representation of the rotation $\mathbf{R}_i$. This is beneficial as over a short period of time there is less scope for error to accumulate in the GPS/IMU measurements. The image measurements $\mathbf{I}^L_i$ and $\mathbf{I}^P_i$, are created using the CNN described in Section~\ref{sec:cnn}, with the priors generated by projecting the skeleton model through the camera using $\mathbf{T}_i$ and the camera intrinsics matrix $\mathbf{K}$.  

To optimise the graph, we define a cost function that computes the residual error between the  expected measurements -- given the current state $\hat{\mathbf{T}_i}$ of the optimiser -- and the sensor measurements described above
\begin{equation}\label{eq:cost-function}
	E = \sum_{i} e^T_{i,i-1} + \sum_{i} e^I_i \enspace.
\end{equation}
The optimal set of poses are those which minimise this function. This function is optimised each time a new pose is added to the graph, and is initialised with the results from the previous optimisation. The first term $e^{T}_{i,i-1}$ compares the relative pose of the current estimates $\hat{\mathbf{T}}_{i,i-1}$ with those obtained from the GPS and IMU, i.e. 
\begin{equation}\label{eq:relative-pose-diff}
	e_{i,i-1}^{T} =  \mathbf{C}
	\begin{bmatrix}
		\hat{\mathbf{t}}_{i,i-1} - \mathbf{t}_{i,i-1} \\[0.5em]
		2 \times \text{Vec} \left(  \hat{\mathbf{q}}_{i,i-1} * \mathbf{q}_{i,i-1}^{-1} \right)
	\end{bmatrix} \enspace,
\end{equation}
where `Vec' corresponds to the vector part of the quaternion rotation, and $\mathbf{C}$ is a diagonal matrix weighting the different elements of the cost.  Due to limitations in our flight software, the covariances of the relative pose measurements are unavailable. Instead we set the values in $\mathbf{C}$ manually prior to the optimisation, based on the expected error in the IMU/GPS measurements, weighting each term in the translation error by $\beta^t$ and those in the rotation error by $\beta^q$. The output $e^{T}_{i,i-1}$ is then a 6D vector containing the residuals in rotation and translation.
\begin{figure}[t]
	\centering
	\begin{tabular}{ccc}
	\includegraphics[width=0.3\linewidth]{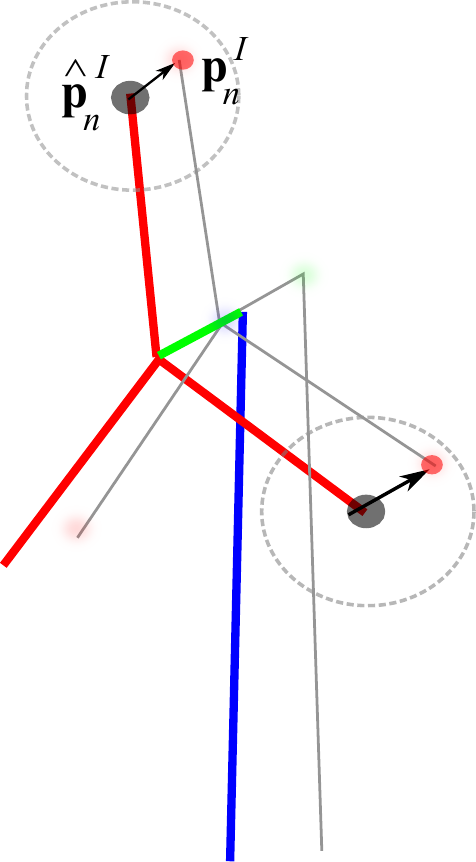}&&\includegraphics[width=0.3\linewidth,]{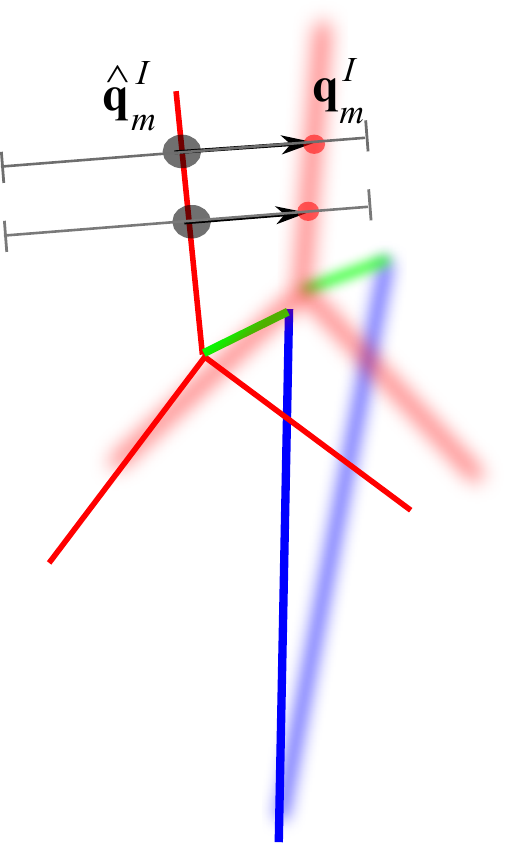}\\[1ex] 			(a)&&(b)
	\end{tabular}
	\caption{Correspondence matching. (a) Point-based correspondences are found by searching within a circular window. (b) Line-based correspondences are found using a perpendicular line search. Model projections inferred by the CNN are shown as blurred points and lines and the model projection using the current pose estimate $\hat{\mathbf T}_i$ is shown in bold colours.}
	\vspace{-1.5em}
	\label{fig:constraint_images}
\end{figure}

The cost function for the image measurements is based on point-to-point correspondences which are established differently depending on the types of image measurements. Figure~\ref{fig:constraint_images} provides an illustration. To establish correspondences we adopt an active search approach as follows. For each of the point-based measurements $\mathbf{I}^P_i$, we project the points $\mathbf{p}_n^c \in \mathcal{P}^c$, using the current estimated pose of the camera $\hat{\mathbf{T}}_i$ and the camera intrinsic matrix $\mathbf{K}$ to find their location on the image plane $\hat{\mathbf{p}}^{I}_n$. Given this location, we extend a circular window of radius $r^P$ and search the pixels within the window to find the one with the largest value which is selected as the correspondence $\mathbf{p}^{I}_n$, as shown in Fig.\ \ref{fig:constraint_images}a. If there are no pixels with value above a threshold $\lambda^P$ then no correspondence is established. This process is repeated for all the points in the turbine model and across all the point image measurements.

\begin{figure*}[t]
	\centering
	\includegraphics[width=0.9\linewidth]{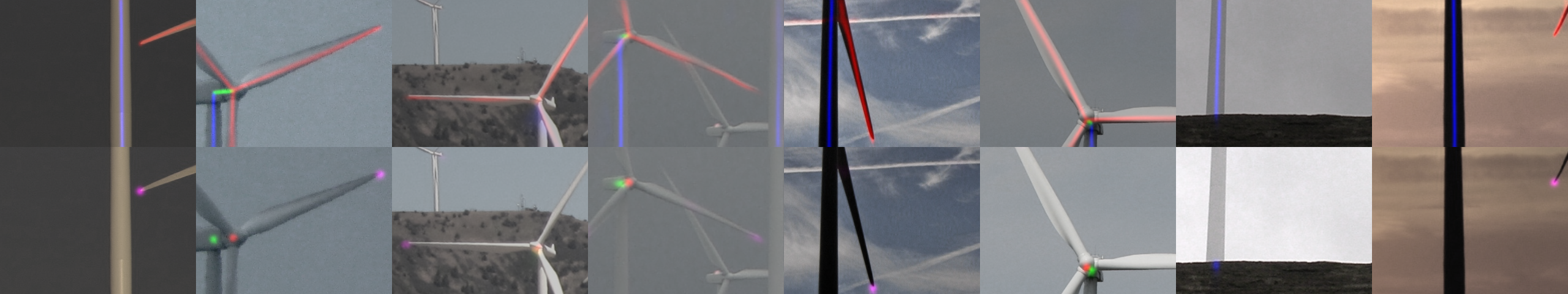}
	\caption{Example output from the CNN. Top) Lines output $\mathbf{I}^L$. Bottom) Points output $\mathbf{I}^P$}
	\vspace{-1em}
	\label{fig:cnn_prior_exp_output1}
\end{figure*}

For the line-based measurements $\mathbf{I}_i^L$, the process is more complicated. First, we extract a set of points $\mathcal{Q}^c = \left\{\mathbf{q}^c_1, \dots, \mathbf{q}^c_m \right\}$ from the lines in $\mathcal{L}^c$. This is done by subdividing the lines at regular intervals. The number of subdivisions for each type of line is different due to their differing lengths. The tower is subdivided into $s^t$ points, the hub into $s^h$ points and the blades into $s^b$ points. Next, we project each of the subdivided points $\mathbf{q}^{c}_m$ into the image using $\hat{\mathbf{T}}_i$ and $\mathbf{K}$ to get their 2D locations $\hat{\mathbf{q}}^I_m$. Instead of the circular search area described for the point-based correspondences, we instead do a perpendicular line search from the projected points as shown in Fig. \ref{fig:constraint_images}b. The reason for this is that often the projected line, and the corresponding line in $\mathbf{I}_i^L$ will be near parallel to one-another. We therefore want the correspondence to be the closest point perpendicular to the projected line.

To perform the perpendicular line search, for each line in $\mathcal{L}^c$, we project the two end points into the image space and find the 2D line connecting them. The 2D direction perpendicular to this line is the direction used during the line search. For each of the subdivided points $\hat{\mathbf{q}}^I_m$ we then sample the pixel values perpendicular to the line over a length of $a^L$ with $k^L$ sample locations. The pixel with the highest value is  chosen as the correspondence $\mathbf{q}^I_m$. Again, if there are no values above a threshold $\lambda^L$, no correspondence is established. After finding the set of correspondences for each frame as described above, we can define the image cost function as
\begin{equation}\label{eq:new-point-residual}
	e^{I}_i = \beta^p \sum_{n}^{N_i} | \hat{\mathbf{p}}^I_n - \mathbf{p}^I_n | + \beta^q \sum_{m}^{M_i} | \hat{\mathbf{q}}^I_m - \mathbf{q}^I_m | \enspace,
\end{equation}
where $N_i$ represents to the number of point correspondences in the frame $i$ and $M_i$ represents to the number of line correspondences in frame $i$. The values $\beta^p$ and $\beta^q$ are used to weight the different types of correspondences. 

With the cost function fully defined we are able to optimise the pose graph. This is done using the Gauss-Newton algorithm which works by linearising the problem around the current best guess solution, finding the minimum and repeating until convergence. As we are provided with initial pose estimates from the GPS / IMU, we found this method appropriate for our problem.

\section{EXPERIMENTS}\label{sec:experiments}

As we do not have access to ground truth pose estimates for the inspection flights, we are unable to perform a quantitative evaluation of the overall performance of the method. However, we are able to evaluate the different sections in isolation. In Section~\ref{sec:eval-network}, we evaluate the performance of the CNN, and show the importance of incorporating prior information into the network input. In Section~\ref{sec:eval-sim} we show the performance of the pose estimation part of the system using synthetic data. Finally, in Section~\ref{sec:eval-real} we give a qualitative evaluation of the system using real-world inspection data.

\subsection{Network Evaluation}\label{sec:eval-network}

\begin{figure}[t]
	\centering
	\includegraphics[width=0.9\linewidth]{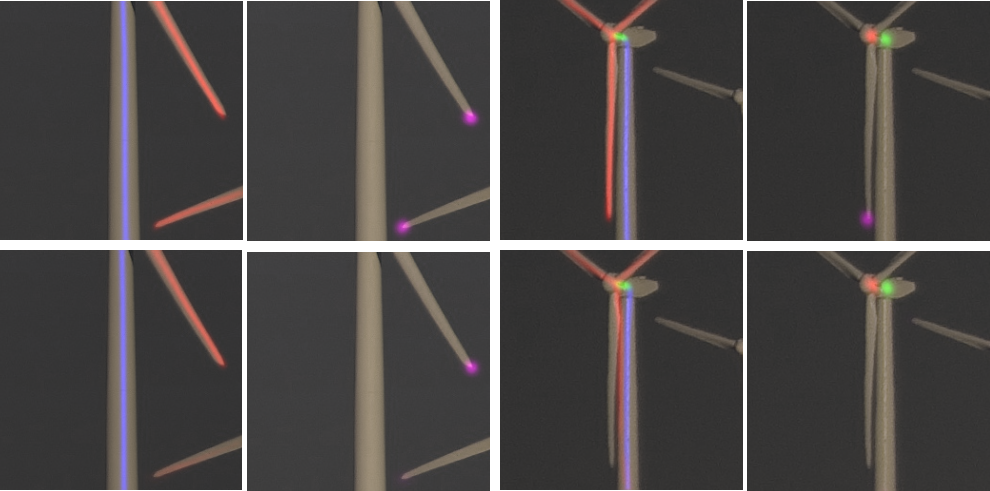}
	\caption{Example line outputs from the evaluation of the effect of using prior in the CNN. Top row) Network with prior information, Bottom row) Network without prior information}
	\vspace{-1em}
	\label{fig:cnn_prior_exp_output2}
\end{figure}

Figure~\ref{fig:cnn_prior_exp_output1} shows example outputs from the CNN for 8 partial views of wind turbines, with the line and point estimates shown in the top and bottom rows, respectively. The colours indicated the different line and point classes. Note that even with a very limited view of the turbine, the network is able to accurately predict the projection of the lines and points of the skeleton model. We also evaluated the impact of including prior information about the turbine lines and points with the input on the CNN performance. We trained two versions of our architecture, one including prior information, and the other without prior information. The networks were identical apart from the shape of the filters in the first layer which enable the inclusion of the extra channels. The networks were trained using identical data sets and for exactly the same number of epochs. To evaluate the performance, we applied the networks to a set of test data and computed the pixel-wise mean squared error between the network predictions and the ground truths. The mean squared error for the network with priors was 0.001, and the mean squared error for the network without priors was 0.0012. This shows that performance is improved by including prior information. This is backed up in Fig.~\ref{fig:cnn_prior_exp_output2} were we compare the two network outputs. We can see that when the prior is included, the performance of the CNN is much more consistent, particularly in predicting the blade lines.

\subsection{Simulated Evaluation}\label{sec:eval-sim}

\begin{figure}[t]
	\centering
	\includegraphics[width=0.75\linewidth]{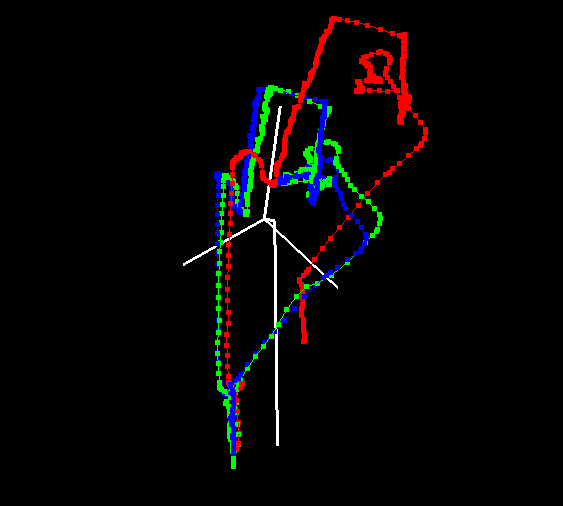}
	\caption{Output from synthetic experiments. Ground truth path is in blue, input path is red, and the path after optimisation is in green.}
	\vspace{-1.5em}
	\label{fig:synth_visual}
\end{figure}

To provide a investigatory evaluation the performance of the pose estimation part of the system, we designed an experiment using synthetic data. We first extracted the GPS/IMU poses from a set of actual inspection flights to give some example flight paths to use as ground truths. We next added progressive Gaussian noise to the ground truths on both the translations and rotations to provide us with an example of the sort of errors we would expect to accumulate over the course of an inspection. For the translations, we sampled a random 3D offset from a zero mean Gaussian distribution and added it to each of the nodes. For the rotation, we sampled a random angle from a zero mean Gaussian distribution, as well as a randomised normalised vector and applied this as an axis-angle rotation to each of the poses. This gave us the simulated IMU/GPS poses which at the start of the flight are close to the ground truth with the error getting progressively worse throughout the inspection. An example can be seen in Figure~\ref{fig:synth_visual}. These noisy poses are used as the relative pose measurements $\mathbf{T}$ in the optimiser. To generate the image measurements, we directly simulate the output of the network by projecting the lines $\mathcal{L}^c$ and points $\mathcal{P}^c$ representations through the set of ground truth poses at each node of the graph. Although the simulated image measurements are error free -- unlike the typical output of the network -- these are suitable for evaluating the performance of the pose optimisation.

\begin{figure}[t]
	\centering
	\includegraphics[width=0.9\linewidth]{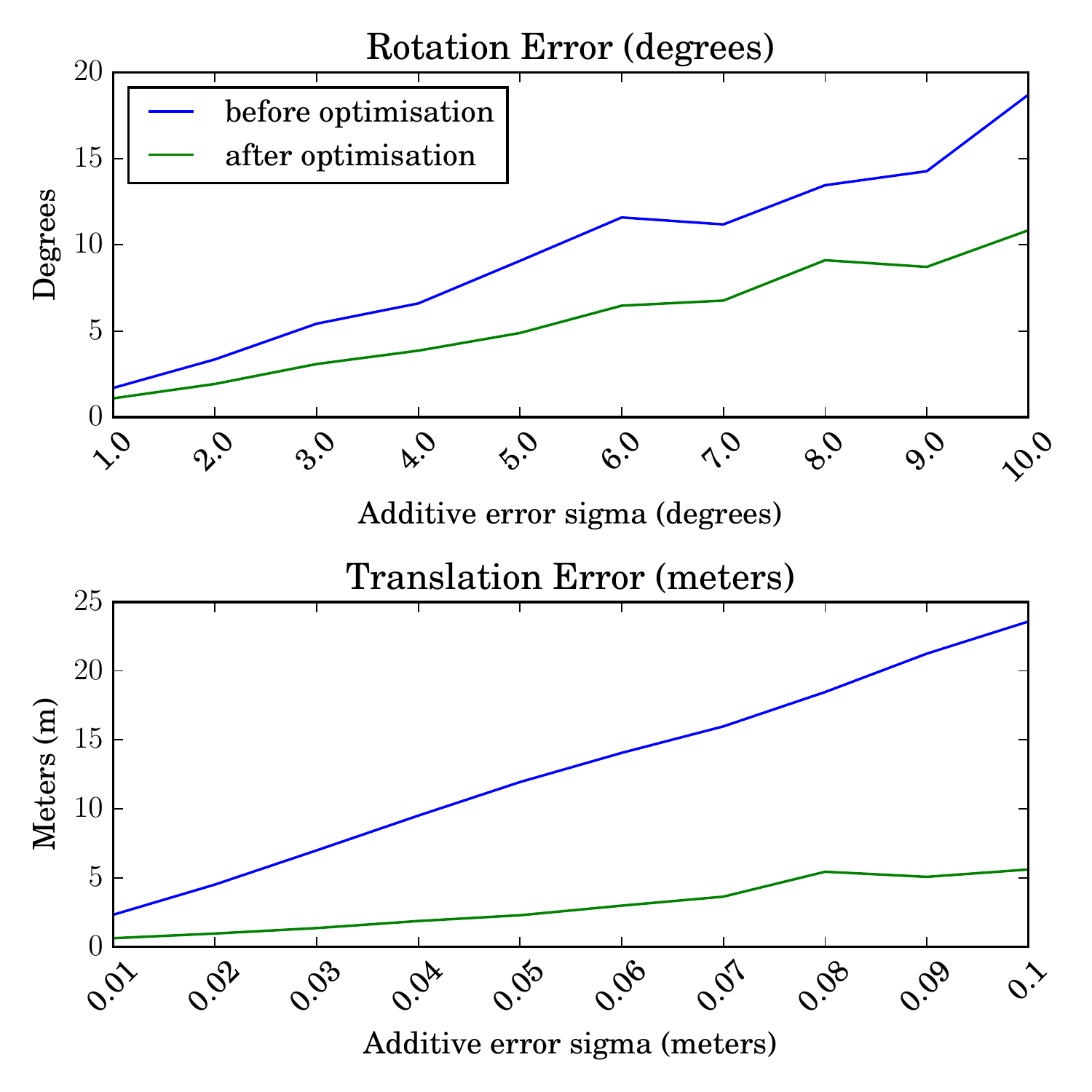}
	\vspace{-2ex}
	\caption{Synthetic experiment results. Top) Pre and post optimisation rotation error. Bottom) Pre and post optimisation translation error.}
	\label{fig:synth_plot}
\end{figure}
The measure that we are interested in evaluating is the robustness of the optimiser as increasing amounts of error are added to the relative pose measurements. Knowing this gives us an understanding of how accurate the initial GPS/IMU pose measurements need to be to be able to reasonably correct the localisation. To evaluate this, we generated a series of synthetic flights as described above with translations errors ranging from 0.01m - 0.1m, and rotation errors from 1-10 degrees. We applied our method to these datasets and recorded the average translation error and rotation error for each flight. The results can be seen in Figure~\ref{fig:synth_plot} and an example output can be seen in Figure~\ref{fig:synth_visual}. From the plots we can see that the system is able to handle quite a large amount of error in the pose measurements, especially for the translation. These preliminary results about the effectiveness of the method are encouraging, although more detailed analysis is needed to fully explore the failure states.

\begin{figure}[t]
	\centering
	\includegraphics[width=0.9\linewidth]{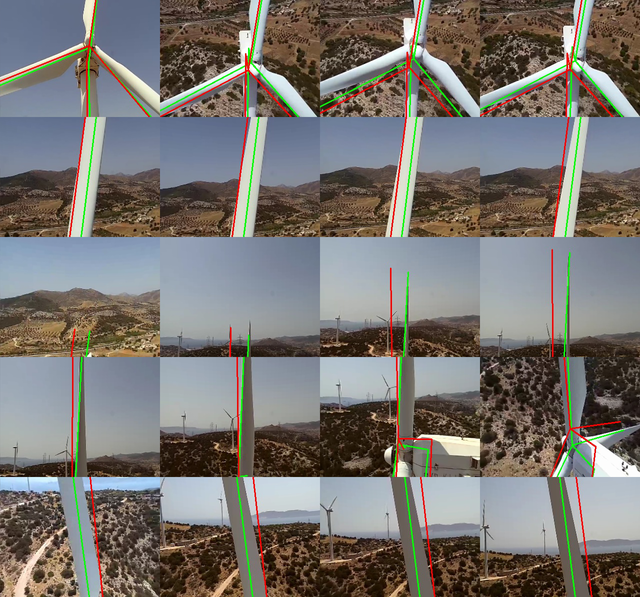}
	\caption{Examples of the method being applied to real inspection data. The turbine model has been projected into the image using the pose prior to optimisation (red)  after optimisation (green).}
	\vspace{-1.5em}
	\label{fig:flight_output}
\end{figure}

\subsection{Visual Evaluation on Real Data}\label{sec:eval-real}

Our final experiment aims to provide a qualitative evaluation of the method applied to a real inspection flight. Although our method was applied after the flight had taken place, it was done in a way that exactly mimics how it would work if it were running during the flight. Examples of the optimiser output can be seen in Figure~\ref{fig:flight_output}. From the images, we can see that the described method noticeably improves the localisation.

\section{CONCLUSIONS}\label{sec:conclusion}

We have presented a novel method for integrating image-based measurements into drone localisation for the  automated inspection of wind turbines. We have described a novel CNN-based system of producing a simplified representation of the wind turbine that allows for easy matching with our wind turbine model. We have also detailed how this representation is incorporated into a pose graph optimisation system. We evaluated the different sections of our work separately. We showed that the inclusion of prior information into the network, improved prediction performance by comparing it to a identical network without prior information. We also evaluated the performance of the pose estimation using synthetic data. Finally we gave a qualitative evaluation of the complete system when applied to an inspection flight.

There are a number of different avenues for future work. First, we would like to properly integrate the GPS/IMU measurements into the system, allowing us to obtain the covariances from these sensors which should improve the optimisation. We also intend to incorporate additional sensors such as LiDAR. Finally, we intend to would look at simultaneously estimating the parameters of the turbine models as part of the optimisation.

\section*{ACKNOWLEDGEMENTS}

The authors acknowledge the support of Innovate UK (project number 104067). The work described herein is the subject of UK patent applications GB1815864.2 and GB1902475.1


\bibliographystyle{IEEEtran}
\bibliography{mybib}


\end{document}